\documentclass[sigconf, screen, table]{acmart}
\usepackage{pifont}
\AtBeginDocument{%
  }

\copyrightyear{2025}
\acmYear{2025}
\setcopyright{acmlicensed}\acmConference[MM '25]{Proceedings of the 33rd ACM International Conference on Multimedia}{October 27--31, 2025}{Dublin, Ireland}
\acmBooktitle{Proceedings of the 33rd ACM International Conference on Multimedia (MM '25), October 27--31, 2025, Dublin, Ireland}
\acmDOI{10.1145/3746027.3758206}
\acmISBN{979-8-4007-2035-2/2025/10}

\usepackage{hyperref} 

\usepackage{pifont}
\usepackage{graphicx}
\usepackage{wrapfig}
\usepackage{graphicx}
\usepackage{booktabs}
\usepackage{svg} % 需使用包
\usepackage{colortbl}  %彩色表格需要加载的宏包

\usepackage{multirow}
\usepackage{indentfirst}
\usepackage{threeparttable}
\usepackage{pifont}
\usepackage{makecell}
\usepackage{arydshln}
\usepackage{arydshln}
\usepackage{tabularx}

\usepackage{makecell}

\usepackage{pifont}
\usepackage{minitoc}
\definecolor{mblue}{RGB}{0, 77, 128}

\definecolor{mblue}{RGB}{0, 77, 128}
\definecolor{mred}{RGB}{192,0, 0}
\definecolor{darkgreen}{rgb}{0.0, 0.5, 0.0}
\definecolor{mycolor_blue}{HTML}{E7EFFA}

\definecolor{mycolor_gray}{HTML}{ECECEC}
\newcommand{\cmark}{\textcolor{darkgreen}{\ding{51}}}  % 深绿色的勾
\newcommand{\xmark}{\textcolor{red}{\ding{55}}}        % 红色的叉
\hypersetup{
    colorlinks=true,        % 如果想要彩色链接
    linkcolor=blue,        % 用于普通链接的颜色
    filecolor=magenta,     % 用于文件链接的颜色
    urlcolor=magenta,          % 用于URL的颜色
}

\begin{document}
\title{3DGS-IEval-15K: A Large-scale Image Quality Evaluation Database for 3D Gaussian-Splatting}
% \author{
% Yuke Xing\textsuperscript{1},
%     Jiarui Wang\textsuperscript{1*}, 
%     Peizhi Niu\textsuperscript{2*},
%     Wenjie Huang\textsuperscript{1},
%     %William Gordon\textsuperscript{1},
%     Guangtao Zhai\textsuperscript{1},
%     Yiling Xu\textsuperscript{1\dag} \\
%     \textsuperscript{1}\{xingyuke-v, wangjiarui, huangwenjie2023, zhaiguangtao, yl.xu\}@sjtu.edu.cn, 
%      \textsuperscript{2}peizhin2@illinois.edu, 
%      \\
%     \textsuperscript{1}{Shanghai Jiao Tong University, Shanghai, China} \\
%     \textsuperscript{2}{University of Illinois Urbana-Champaign, Urbana, USA} \\
%     \textsuperscript{*}Equal contribution
%     \textsuperscript{\dag}Corresponding author
%     \\
% }

\author{Yuke Xing}
\email{xingyuke-v@sjtu.edu.cn}
\affiliation{%
  \institution{Shanghai Jiao Tong University}
  \city{Shanghai}
  \country{China}
}

\author{Jiarui Wang}
\authornote{Equal Contribution.}
\email{wangjiarui@sjtu.edu.cn}
\affiliation{%
  \institution{Shanghai Jiao Tong University}
  \city{Shanghai}
  \country{China}
}

\author{Peizhi Niu}
\authornotemark[1]
\email{peizhin2@illinois.edu}
\affiliation{%
  \institution{University of Illinois Urbana-Champaign}
  \city{Urbana}
  \country{USA}
}

\author{Wenjie Huang}
\email{huangwenjie2023@sjtu.edu.cn}
\affiliation{%
  \institution{Shanghai Jiao Tong University}
  \city{Shanghai}
  \country{China}
}

\author{Guangtao Zhai}
\email{zhaiguangtao@sjtu.edu.cn}
\affiliation{%
  \institution{Shanghai Jiao Tong University}
  \city{Shanghai}
  \country{China}
}
\author{Yiling Xu}
\authornote{Corresponding author.}
\email{yl.xu@sjtu.edu.cn}
\affiliation{%
  \institution{Shanghai Jiao Tong University}
  \city{Shanghai}
  \country{China}
}

\renewcommand{\shortauthors}{Yuke Xing, et al.}

\begin{abstract}
 3D Gaussian Splatting (3DGS) has emerged as a promising approach for novel view synthesis, offering real-time rendering with high visual fidelity. However, its substantial storage requirements present significant challenges for practical applications. 
 While recent state-of-the-art (SOTA) 3DGS methods increasingly incorporate dedicated compression modules, there is a lack of a comprehensive framework to evaluate their perceptual impact. Therefore we present \textbf{3DGS-IEval-15K}, the first large-scale image quality assessment (IQA) dataset specifically designed for compressed 3DGS representations. Our dataset encompasses 15,200 images rendered from 10 real-world scenes through 6 representative 3DGS algorithms at 20 strategically selected viewpoints, with different compression levels leading to various distortion effects. Through controlled subjective experiments, we collect human perception data from 60 viewers. We validate dataset quality through scene diversity and MOS distribution analysis, and establish a comprehensive benchmark with 30 representative IQA metrics covering diverse types. As the largest-scale 3DGS quality assessment dataset to date, our work provides a foundation for developing 3DGS specialized IQA metrics, and offers essential data for investigating view-dependent quality distribution patterns unique to 3DGS. The database is publicly available
at \url{https://github.com/YukeXing/3DGS-IEval-15K}.
\end{abstract}





\definecolor{mycolor_green}{HTML}{E6F8E0}
\maketitle

%\input{figures/source_content}
%\vspace{-10pt} 
\section{Introduction}

\begin{figure*}[!t]
    \centering
    \vspace{-4mm}
    \includegraphics[width=0.98\linewidth]{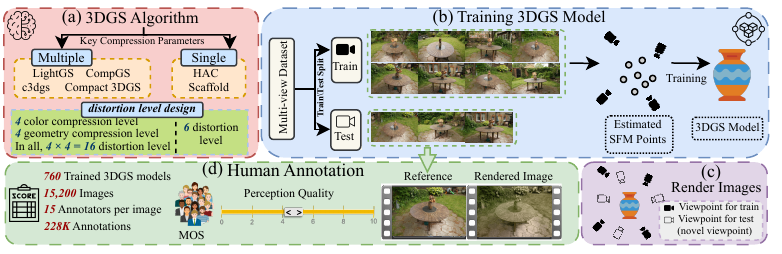}
    \vspace{-4mm}
    \caption{
         Overview of the construction pipeline of 3DGS-IEval-15K, designed for benchmarking 3DGS representations. (a) We select 6 mainstream compressed 3DGS algorithms, including 4 multi-compression-parameter methods designed with 4 geometry and color compression levels each (combined to produce 16 distinct distortion levels), and 2 single-compression-parameter methods designed with 6 distortion levels each. (Section~\ref{3DGS_Model_and_Bitrate_Point_Selection}) (b) We select 10 scenes from multi-view datasets, each containing numerous viewpoints. For each scene, we split the viewpoints into training/testing set and use the training set to reconstruct 3D scenes. (Section~\ref{Source_Content_Selection}) (c) For each scene, we select 10 representative viewpoints from the training set and 10 challenging viewpoints from the testing set as defined in (b). Reconstructed 3D scenes are then rendered from these 20 viewpoints. (Section~\ref{3D_Viewpoint_Selection}) (d) With 15 annotations per image, we create a large-scale 3DGS image quality evaluation dataset (3DGS-IEval-15K) with labeled MOS scores. (Section~\ref{Subjective_Experiment_and_Data_Processing})
    }
    \label{overview}
    \vspace{-3mm}
\end{figure*}

Neural View Synthesis (NVS) has emerged as a transformative technology in computer vision, enabling the generation of photorealistic images from arbitrary camera viewpoints given sparse input views. The field has witnessed remarkable progress with the introduction of Neural Radiance Fields (NeRF) ~\citep{nerf} and its subsequent evolution into 3DGS ~\citep{3DGS} , which has revolutionized real-time rendering capabilities while maintaining high visual fidelity. However, 3DGS demands substantially more storage resources, significantly hindering practical deployment. Consequently, 3DGS compression has become critical, with recent SOTA 3DGS algorithms ~\citep{Scaffold-GS,HAC,Lightgaussian,CompGS,c3dgs,Compact-3DGS,eagles,liu2024compgs} increasingly incorporating compression modules. 
Effective quality assessment (QA) metrics are essential for guiding 3DGS training and identifying optimal compression strategies. However, existing research predominantly relies on traditional IQA metrics, which may overlook unique 3DGS distortions resulting in inaccurate prediction. This necessitates a comprehensive dataset with diverse 3DGS distortion effects for designing and evaluating specialized QA metrics.

Current NVS-QA datasets ~\citep{NeRF-QA, NeRF-VSQA,FFV,ENeRF-QA,GSC-QA,GS-QA,NVS-QA} exhibit several critical limitations: (1) \textbf{Lack of distortion sample design incorporating NVS compression.} Despite compression being a research priority, existing datasets fail to systematically incorporate diverse compression parameters, resulting in inadequate distortion variety for effective metric training.
(2) \textbf{Severely constrained scales.} Due to substantial training costs and per-scene requirements, most datasets contain fewer than 100 samples and even the largest not exceeding 500, insufficient for training robust objective models. 
(3) \textbf{Lack of attention to 3DGS image quality assessment.} While video quality assessment (VQA) simulates continuous viewpoint exploration, image quality assessment (IQA) is crucial for efficient training feedback and investigating view-dependent quality variations unique to 3DGS. However, existing datasets predominantly focus on VQA, leaving 3DGS-IQA as a significant research gap.

To address these limitations, we present \textbf{3DGS-IEval-15K}, the first large-scale image quality assessment dataset specifically designed for compressed 3DGS representations. Our dataset encompasses 15,200 images rendered from 10 real-world scenes through 6 mainstream 3DGS compression algorithms at 20 carefully selected viewpoints, with different compression levels leading to various distortion effects, providing unprecedented scale and systematic coverage of compression methods and distortion types. We establish a comprehensive benchmark encompassing 30 representative IQA metrics, systematically including deep learning-based and large language model (LLM)-based approaches for the first time on 3DGS database, providing comprehensive evaluation capabilities that were previously unavailable due to insufficient dataset scale. Our viewpoint selection strategy identifies both representative training perspectives and challenging test perspectives with maximal differences from the training set, enabling fine-grained analysis of view-dependent quality variations unique to 3DGS.
The main contributions of our work are summarized as follows:

$\cdot$ \textbf{largest 3DGS IQA dataset}: We generate 15,200 samples with systematic compression-based distortion design, enabling specialized 3DGS quality assessment model training and facilitating 3DGS training processes optimization.

$\cdot$ \textbf{First comprehensive 3DGS-IQA benchmark}: We  systematically evaluate 30 objective quality metrics, including deep learning and LLM-based approaches, revealing the limitations of existing methods and providing suggestions for metric selection and design.

$\cdot$ \textbf{View-dependent quality analysis foundation}: We provide essential data for investigating unique 3DGS quality distribution patterns, offering insights into viewpoint-dependent reconstruction fidelity and suggests potential optimization strategies align with human visual perception.

\begin{table*}[!]
\centering
\vspace{-3mm}
\caption{Summary for existing NVS quality evaluation datasets, with ``Syn'' and ``Real'' representing synthetic and real scenes. }
\vspace{-3mm}
\renewcommand\arraystretch{0.85}
\label{tab:relate}
\resizebox{0.95\textwidth}{!}{\begin{tabular}{ccccccccccccc}
\hline
Dataset & \multirow{2}{*}{Name} & \multirow{2}{*}{Year}  
&\multicolumn{2}{c}{Scenes} & 
\multicolumn{2}{c}{NVS Models} 
& Render
& Distortion 
&\multirow{2}{*}{Annotation}
& Number of & Number of
\\

 Type & & & Real & Syn & NeRF & 3DGS & Mode & Level design & & NeRF/3DGS &videos/imgs \\
 \hline

\multirow{7}{*}{VQA} & NeRF-QA~\citep{NeRF-QA}  & 2023 & 4 & 4 & 7 & - &$360^{\circ}$ & \xmark & DMOS & 48 & 48\\
& NeRF-VSQA~\citep{NeRF-VSQA} & 2024 & 8 & 8  & 7 & - & $360^{\circ}$+Front & \xmark & DMOS & 88 & 88 \\
& FFV~\citep{FFV} & 2024 & 22 & - & 8 & - & \textit{Front} & \xmark & Pairs & 220 & 220 \\
& ENeRF-QA~\citep{ENeRF-QA} & 2024 & - & 22 &  4 & - & $360^{\circ}$ & \cmark & MOS & 440 & 440 \\
& GSC-QA~\citep{GSC-QA} & 2024 & 6 & 9 & - & 1 & $360^{\circ}$ & \cmark & MOS & 120 & 120 \\
& GS-QA ~\citep{GS-QA}& 2025 & 8 & - & - & 7 & $360^{\circ}$ + Front & \xmark & DMOS & 64 & 64 \\
& NVS-QA~\citep{NVS-QA} & 2025 & 13 & - & 2 & 3 & $360^{\circ}$+Front & \xmark & MOS & 65 & 65 \\
%\hdashline
\cdashline{1-12}[7pt/5pt]  % 长度4pt，间隔3pt的虚线，横跨1-7列
\multirow{2}{*}{IQA} & NVS-QA~\citep{NVS-QA} & 2025 & 13 & - & 2 & 3 & 1 viewpoints& \xmark & MOS & 65 & 65 \\
& \cellcolor{gray!20}\textbf{3DGS-IEval-15K(Ours)} 
& \cellcolor{gray!20}- 
& \cellcolor{gray!20}\textbf{10} 
& \cellcolor{gray!20}- 
& \cellcolor{gray!20}- 
& \cellcolor{gray!20}\textbf{6} 
& \cellcolor{gray!20}\textbf{20 viewpoints}
& \cellcolor{gray!20}\cmark 
& \cellcolor{gray!20}\textbf{MOS} 
& \cellcolor{gray!20}\textbf{760} & \cellcolor{gray!20}\textbf{15,200} \\
\hline
\end{tabular}}
\vspace{-2mm}
\end{table*}

% \begin{itemize}
%     \item First, we construct the largest IQA dataset for 3DGS, featuring 15,200 samples with systematic compression-based distortion design. This enables the training of specialized 3DGS quality assessment models and facilitates the optimization of 3DGS generation processes.
    
%     \item Second, we establish the first comprehensive benchmark for 3DGS image quality assessment, systematically evaluating \textbf{32} objective quality metrics, including deep learning-based and LLM-based approaches, on compressed 3DGS content. This reveals the limitations of existing methods and provides suggestions for metric selection as well as guidance for future metric development.
    
%     \item Third, we provide essential data and foundations for investigating view-dependent quality distribution patterns unique to 3DGS. This offers insights into viewpoint-dependent reconstruction fidelity and suggests potential optimization strategies for 3DGS compression techniques that better align with human visual perception.
% \end{itemize}

\begin{figure*}[!t]
    \centering
    \vspace{-1mm}
    \includegraphics[width=0.95\linewidth]{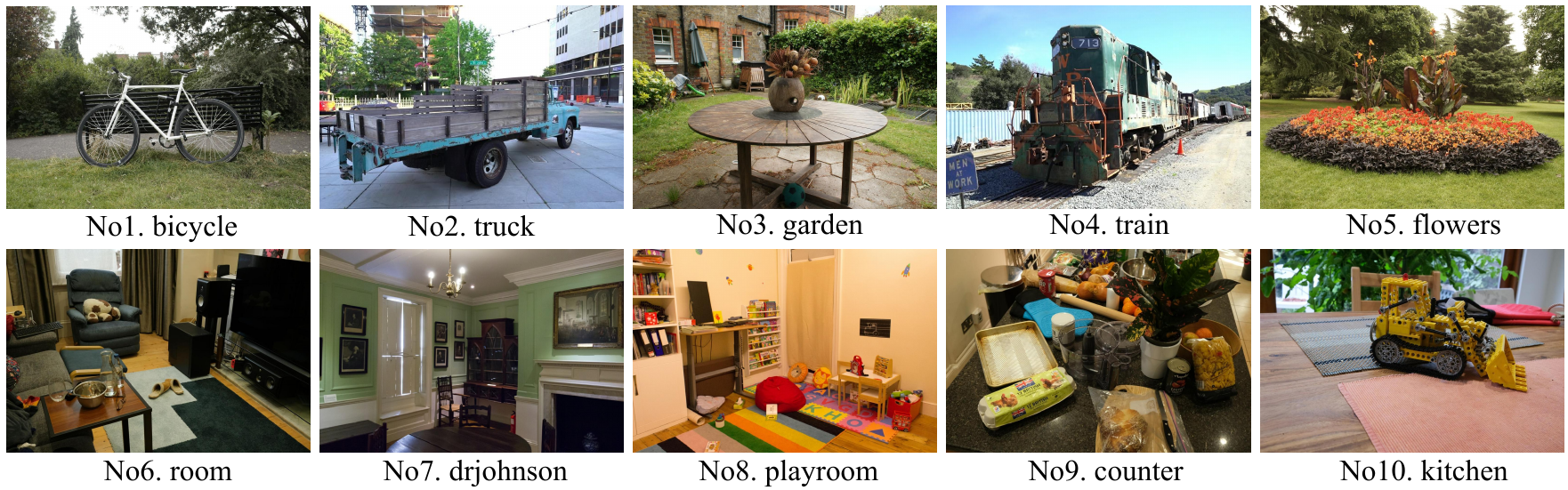}
    \vspace{-3mm}
    \caption{10 selected source content in 3DGS-IEval-15K: scenes 1-5 depict outdoor scenes, while scenes 6-10 depict indoor scenes.} 
    \label{figure:model}
    \vspace{-3mm}
\end{figure*}
\vspace{-2mm}
% \begin{figure*}[!t]
%     \centering
%     % \vspace{-2mm}
%     \includegraphics[width=0.95\linewidth]{figures/train_test_selection.pdf} 
%     \label{fig:train_test_selection}
%     \vspace{-2mm}
%     \caption{ Illustration of the proposed viewpoints selection strategy: Training viewpoints are selected through feature-based $k$-means clustering, while testing viewpoints are chosen based on four criteria.}
%     \label{fig:selection_combined}
%     \vspace{-2mm}
% \end{figure*}

\section {Related Work}

As summarized in Table~\ref{tab:relate}, NVS quality assessment has evolved alongside NVS technologies, transitioning from NeRF-based to 3DGS-focused evaluation. Early NeRF datasets established foundational frameworks: NeRF-QA ~\citep{NeRF-QA} and NeRF-VSQA ~\citep{NeRF-VSQA} provided initial benchmarks with 48 and 88 video samples, while FFV ~\citep{FFV} pioneered pairwise comparison with 220 samples. ENeRF-QA ~\citep{ENeRF-QA} introduced systematic distortion design through NeRF compression across 440 samples, with identifying nine NeRF-specific distortion types.
As 3DGS emerged as superior NVS technology, evaluation focus shifted accordingly. GSC-QA ~\citep{GSC-QA} examined compression effects of their 3DGS compression method with 120 samples, GS-QA ~\citep{GS-QA} compared 3DGS and NeRF methods across 64 samples, while NVS-QA ~\citep{NVS-QA} pioneered both video and image assessment with 65 samples each.

However, existing datasets face critical limitations hindering comprehensive evaluation. First, scales remain severely constrained with most containing fewer than 100 samples and even the largest not exceeding 500, insufficient for training robust objective models. Second, most lack systematic compression integration for distortion design, despite compression being essential for 3DGS practical deployment due to substantial storage requirements. Third, the predominant video-focus overlooks image quality assessment's irreplaceable importance for 3DGS generation, where IQA enables efficient training feedback and investigation of view-dependent quality variations unique to 3DGS.

To address these limitations, we propose 3DGS-IEval-15K, the first large-scale IQA benchmark specifically designed for 3DGS evaluation. It contains 15,200 samples featuring systematically constructed compression-induced distortions, enabling comprehensive assessment of representative compressed 3DGS methods.

\vspace{-1mm}
\section{DATABASE CONSTRUCTION}

\begin{figure*}[!t]
    \centering
     \vspace{-2mm}
    \includegraphics[width=0.8\linewidth]{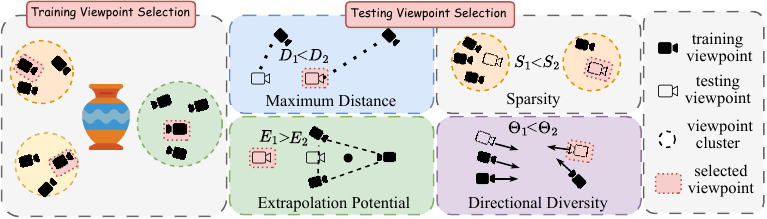} 
    \label{fig:train_test_selection}
    \vspace{-2mm}
    \caption{ Illustration of the proposed viewpoints selection strategy: Training viewpoints are selected through feature-based $k$-means clustering, while testing viewpoints are chosen based on four criteria.}
    \label{fig:selection_combined}
    \vspace{-2mm}
\end{figure*}

\subsection{Source Content Selection}
\label{Source_Content_Selection}
Our dataset comprises 10 real-world scenes selected from three canonical multiview datasets in the NVS domain to ensure diverse visual characteristics for comprehensive quality assessment. As illustrated in Figure~\ref{figure:model}, the collection includes six scenes from \textit{Mip-NeRF 360}~\citep{Mip-NeRF-360}, consisting of three outdoor scenes: \textit{bicycle} with resolution 1237~$\times$~822, \textit{flowers} with resolution 1256~$\times$~828, \textit{garden} with resolution 1297~$\times$~840, and three indoor scenes: \textit{counter} with resolution 1558~$\times$~1038, \textit{kitchen} with resolution 1558~$\times$~1039, \textit{room} with resolution 1557\allowbreak{}~$\times$~1038. Additionally, two outdoor scenes are picked from \textit{Tanks \& Temples}~\citep{Tanks-Temples}: \textit{train} with resolution 980~$\times$~545, \textit{truck} with resolution 979~$\times$~546, and two indoor scenes are picked from \textit{Deep Blending}~\citep{Deep-Blending}: \textit{playroom} with resolution 1264~$\times$~832, \textit{drjohnson} with resolution 1332~$\times$~876. For each scene, we partition the viewpoints into disjoint training and testing set. The training set is then used to reconstruct a 3DGS model.

This selection presents varied challenges for 3DGS representation and compression, including both indoor and outdoor scenes, complex occlusions, different lighting conditions, reflective surfaces, and natural textures with high-frequency details. The resolution settings of each scene in the original 3DGS are maintained, thereby standardizing evaluation protocol and establishing a solid foundation for cross-study comparison and communication.

\vspace{-3mm}
\subsection{3D Viewpoint Selection}
\label{3D_Viewpoint_Selection}

Viewpoint selection is critical for effective evaluation of 3DGS models. We propose a systematic strategy (Figure~\ref{fig:selection_combined}) that selects \textbf{10} representative viewpoints from training set and \textbf{10} challenging viewpoints from testing set mentioned in Section~\ref{Source_Content_Selection}, enabling robust assessment of view-dependent generalization and perceptual quality under diverse conditions.

\textbf{Training Viewpoint Selection:}  To select training viewpoints, we adopt a feature-based clustering strategy that ensures broad scene coverage while minimizing redundancy. For each scene, all the viewpoints from training set is encoded by a feature vector $\mathbf{f} = [\mathbf{p}, \beta\mathbf{d}]$, where $\mathbf{p}$ is the position, $\mathbf{d}$ is the viewing direction, and $\beta = 0.3$ balances their relative importance. These features are then normalized and partitioned using $k$-means clustering, resulting in \textbf{10} distinct clusters. From each cluster, we select the viewpoint closest to the centroid, ensuring that the chosen viewpoints optimally represent the distribution of available camera poses while maintaining diversity.
% :
% \begin{equation}
% \hat{\mathbf{f}}_i = \frac{\mathbf{f}_i - \boldsymbol{\mu}_f}{\boldsymbol{\sigma}_f}
% \end{equation}
% where $\boldsymbol{\mu}_f$ and $\boldsymbol{\sigma}_f$ are the mean and standard deviation of the feature set
% \begin{figure*}[!t]
%     \centering
%     \vspace{-2mm}
%     \includegraphics[width=0.9\linewidth]{figures/train_test_selection.pdf} 
%     \label{fig:train_test_selection}
%     \vspace{-2mm}
%     \caption{ Illustration of the proposed viewpoints selection strategy: Training viewpoints are selected through feature-based $k$-means clustering, while testing viewpoints are chosen based on four criteria.}
%     \label{fig:selection_combined}
%     \vspace{-2mm}
% \end{figure*}

% \input{tables/PL1}
% \input{tables/PL2}

\textbf{Testing Viewpoint Selection:} To rigorously assess the generalization capability of NVS, we select testing viewpoints that exhibit maximal distributional divergence from the viewpoints from training set. Each candidate is ranked using a composite score that integrates multiple criteria. The composite score for a viewpoint $j$ from testing set is formulated as:
\vspace{-5pt}
\begin{equation}
S_j = w_d D_j + w_s S_j + w_e E_j + w_\theta \Theta_j
\vspace{-5pt} 
\end{equation}
where $D_j$ represents normalized distance metrics, $S_j$ quantifies local sparsity, $E_j$ indicates extrapolation requirements, $\Theta_j$ measures directional novelty, and $w$ factors are the respective weights (we set all $w$ values to 0.25). The detailed definitions of these four criteria are as follows:

\textit{Maximum Distance}: Preference for viewpoints maximally distant from any training viewpoints:
\vspace{-5pt} 
\begin{equation}
D_j = \frac{1}{2}\left(\frac{\min_{i \in T} \|\mathbf{p}_j - \mathbf{p}_i\|}{\max_k \min_{i \in T} \|\mathbf{p}_k - \mathbf{p}_i\|} + \frac{\frac{1}{|T|}\sum_{i \in T} \|\mathbf{p}_j - \mathbf{p}_i\|}{\max_k \frac{1}{|T|}\sum_{i \in T} \|\mathbf{p}_k - \mathbf{p}_i\|}\right)
\end{equation}

\textit{Sparsity}: Prioritization of viewpoints in regions with low training viewpoints density:
\vspace{-5pt} 
\begin{equation}
S_j = 1 - \frac{\rho_j - \min_k \rho_k}{\max_k \rho_k - \min_k \rho_k}, \quad \text{where} \quad \rho_j = \frac{1}{\frac{1}{K}\sum_{l=1}^K d_{j,l} + \epsilon}
\vspace{-5pt} 
\end{equation}

\textit{Extrapolation Potential}: Selection of viewpoints outside the convex hull of training viewpoints, necessitating extrapolation rather than interpolation:
\vspace{-5pt} 
\begin{equation}
E_j = \begin{cases}
1 & \text{if } \mathbf{p}_j \notin \text{ConvexHull}(\{\mathbf{p}_i\}_{i \in T}) \\
0 & \text{otherwise}
\end{cases}
\vspace{-5pt} 
\end{equation}

\textit{Directional Diversity}: Emphasis on viewing directions substantially different from training viewpoints:
\vspace{-5pt} 
\begin{equation}
\Theta_j = \frac{\max_{i \in T} \arccos(\mathbf{d}_j \cdot \mathbf{d}_i) - \min_k \max_{i \in T} \arccos(\mathbf{d}_k \cdot \mathbf{d}_i)}{\max_k \max_{i \in T} \arccos(\mathbf{d}_k \cdot \mathbf{d}_i) - \min_k \max_{i \in T} \arccos(\mathbf{d}_k \cdot \mathbf{d}_i)}
\end{equation}

where $T$ is the indices of viewpoints from training set, $p_j$ and $\mathbf{d}_j$ are the position and viewing direction of viewpoint $j$, $k$ ranges over all viewpoints from testing set, $d_{j,l}$ is the distance between viewpoint $j$ to its $l$-th nearest viewpoints from training set, $K$ is the number of viewpoint $j$'s nearest viewpoints from training set (we set $K=min(10, |T|)$), $\epsilon$ is a small constant to prevent division by zero. We compute the composite scores for all viewpoints in the testing set and select the top \textbf{10} with the highest scores.

This targeted selection strategy provides more rigorous evaluation than random sampling by emphasizing viewpoints requiring significant interpolation or extrapolation from training data, enabling reliable assessment of model generalization in challenging novel view synthesis scenarios.

\vspace{-2pt} 
\subsection{3DGS Model and Bitrate Point Selection}
\label{3DGS_Model_and_Bitrate_Point_Selection}
\subsubsection{3DGS Model Selection} \ 

3DGS is a novel view synthesis approach that represents scenes using a collection of learnable 3D Gaussians. Each Gaussian stores two categories of attributes: geometric properties including position $\mu \in \mathbb{R}^3$, opacity $\alpha \in \mathbb{R}$, covariance matrix $\Sigma$ (parameterized scale $s \in \mathbb{R}^3$ and rotation parameters), and view-dependent color properties represented by spherical harmonics (SH) coefficients. 

Despite superior visual quality, 3DGS requires substantial storage resources, making compression essential for practical deployment. Consequently, 3DGS methods increasingly incorporate dedicated compression modules. In this study, we select six such representative 3DGS algorithms to construct our comprehensive evaluation dataset: \textbf{LightGS}~\citep{Lightgaussian} adopts a pruning-then-quantization approach, removing insignificant Gaussians (controlled by \textit{prune\_percents}) and compressing remaining color coefficients via vector quantization (controlled by \textit{vq\_ratio} and \textit{codebook\_size}). \textbf{c3dgs}~\citep{c3dgs} uses intelligent parameter grouping through $k$-means clustering, where \textit{codebook\_size} determines compression strength and \textit{importance\_include} thresholds filter parameters based on their visual contribution. \textbf{Compact 3DGS}~\citep{Compact-3DGS} replaces color representation with hash-grid neural networks (controlled by \textit{hashmap} size) and applies multi-level vector quantization to geometry (controlled by \textit{codebook\_size} and \textit{rvq\_num} depth), achieving over 25$\times$ storage reduction. \textbf{CompGS}~\citep{CompGS} applies straightforward vector quantization to both geometry and color attributes, with compression level solely controlled by \textit{codebook\_size}. \textbf{HAC}~\citep{HAC} employs context-aware compression by modeling spatial relationships between anchors, with \textit{lambda} parameter balancing compression rate and visual quality through adaptive quantization. \textbf{Scaffold}~\citep{Scaffold-GS} fundamentally changes the representation by using anchor points to generate local Gaussians dynamically, where \textit{vsize} parameter controls the spatial resolution of anchor placement. 

\vspace{-1pt} 
\subsubsection{Compression Parameter Level Design}  \

\begin{table*}[tbph]
\centering
\vspace{-2mm}
\renewcommand{\arraystretch}{0.9}
\caption{
4 CL designed separately for color-distortion-controlled compression parameters and geometry-distortion-controlled compression parameters for multi-compression-parameter methods, with 16 DL 3DGS models each generated through pairwise combinations of these parameters with various CL.} 
\vspace{-3mm}
\resizebox{1\linewidth}{!}{
\begin{tabular}{c!{\vrule}c:c!{\vrule}c:c!{\vrule}c:c!{\vrule}c:c}
    \toprule[1pt]
    \textbf{3DGS Model} & \multicolumn{2}{c!{\vrule}}{\textbf{LightGS}} & \multicolumn{2}{c!{\vrule}}{\textbf{c3dgs}} & \multicolumn{2}{c!{\vrule}}{\textbf{Compact 3DGS}} & \multicolumn{2}{c}{\textbf{CompGS}} \\
    \midrule
    \textbf{Distortion Type} & \textbf{Color} & \textbf{Geometry} & \textbf{Color} & \textbf{Geometry} & \textbf{Color} & \textbf{Geometry} & \textbf{Color} & \textbf{Geometry} \\
    \midrule
    \textbf{Key} & vq\_ratio & \multirow{2}{*}{prune\_percents} & codebook\_size & codebook\_size & \multirow{2}{*}{hashmap} & codebook\_size & \multirow{2}{*}{codebook\_size} & \multirow{2}{*}{codebook\_size} \\
    \textbf{Parameters}& codebook\_size && importance\_include & importance\_include && rvq\_num &&\\
    \midrule
    \textbf{CL01} & (1, $2^1$) & 0.95 & (2, 0.6) & (1, 0.3) & 2 & ($2^2$, 1) & $2^1$ & 1 \\
    \textbf{CL02} & (1, $2^2$) & 0.90 & ($2^2$, 0.6) & ($2^2$, 0.3) & $2^2$ & ($2^4$, 1) & $2^2$ & 2 \\
    \textbf{CL03} & (1, $2^5$) & 0.85 & ($2^5$, 0.6) & ($2^4$, 0.3) & $2^9$ & ($2^2$, 6) & $2^3$ & $2^3$ \\
    \textbf{CL04} & (0.6, $2^{13}$) & 0.66 & ($2^{12}$, $0.6 \times 10^{-6}$) & ($2^{12}$, $0.3 \times 10^{-5}$) & $2^{19}$ & ($2^6$, 6) & $2^{12}$ & $2^{12}$ \\
    \bottomrule
\end{tabular}
}
\label{tab:compression_params1}
\end{table*}

\begin{table}[t]
 \vspace{-2mm}
  \caption{6 CL designed for the key parameter of single-compression-parameter methods, generating 6 DL of 3DGS models each accordingly.}
 
  \renewcommand{\arraystretch}{0.9}
  \label{tab:compression_params2}
  \centering
  \vspace{-3mm}
  \resizebox{0.47\textwidth}{!}{
  \begin{tabular}{l!{\vrule}c!{\vrule}cccccc}
    \toprule
    \textbf{3DGS Model} & \textbf{Parameter} & \textbf{DL01} & \textbf{DL02} & \textbf{DL03} & \textbf{DL04} & \textbf{DL05} & \textbf{DL06}
    \\
    \midrule
    \textbf{HAC} & lmbda  & 0.400 & 0.300 & 0.200 & 0.120 & 0.060 & 0.004
    \\
    \textbf{Scaffold} & vsize & 0.250 & 0.200 & 0.160 & 0.120 & 0.080 & 0.001
    \\
    \midrule

  \end{tabular}}
\vspace{-5mm}
\end{table}

Through analysis of these 6 mainstream algorithms and their compression strategies and parameters, we find that their distortion effects can be broadly categorized into controlling \textbf{geometric distortion} or \textbf{color distortion}. Based on the underlying principles and compression parameter types, we categorize the selected 3DGS methods into two types:
(1) \textbf{Single-compression-parameter algorithms} are primarily \textbf{anchor-based} methods controlled by one key compression parameter that mainly affects geometric distortion, such as HAC and Scaffold-GS.
(2) \textbf{Multi-compression-parameter algorithms} directly \textbf{compress different 3DGS attributes with multiple compression strategies and parameters}, affecting both geometry and color distortions, including LightGS, c3dgs, Compact-3DGS, and CompGS.
To better study the distortion types and degrees introduced by different compression parameters and algorithms, we design different \textbf{Compression Levels (CL)} for key compression parameters that affect geometry and color separately. Through pairwise combinations of these CLs, we ultimately obtain different 3DGS \textbf{Distortion Levels (DL)} and types.

For multi-compression-parameter algorithms, we design \textbf{4 CL} for key parameters controlling geometry distortion and color distortion separately, as shown in Table~\ref{tab:compression_params1}, generating $\mathbf{4 \times 4 = 16}$ \textbf{DL} through pairwise combinations.
For single-compression-parameter algorithms, we design \textbf{6 CL} for their key compression parameter as shown in Table~\ref{tab:compression_params2}, corresponding to $\mathbf{6}$ \textbf{DL} to cover a wide quality range.
In total, we train $\mathbf{10\ \text{scenes} \times (4 \times 16 + 2 \times 6)\ \text{DLs} = 720}$ \textbf{3DGS models}, and obtain $\mathbf{720\ \text{3DGS} \times 20\ \text{viewpoints} = 15,\!200}$ \textbf{distorted 3DGS images}.

\subsection{Subjective Experiment and Data Processing}
\label{Subjective_Experiment_and_Data_Processing}
To evaluate the quality of the images in the GS-IQA, we utilize a double stimulus impairment scale, with reference and 3DGS synthesized images
displayed side-by-side. For the MOS annotation type, we use an 11-
level impairment scale proposed by ITU-TP.910 ~\citep{ITU1(siti)}. The images are displayed using an interface designed with Python Tkinter, as illustrated in Figure ~\ref{overview}(d). The experiment was carried out using a 27-inch AOC Q2790PQ monitor in an indoor laboratory environment under standard lighting conditions. To prevent visual fatigue caused by too long experiment time, 15,200 images are randomly divided into 8 smaller groups. A total of 60 students with different background participate in the experiment.

ITUR BT.500 \citep{ITU2} is applied to conduct the outlier detection and subject rejection. The score rejection rate is 2\%. Finally, by averaging the reserved subjective scores, we obtain MOS score of each image.

\section{Experiments}

\subsection{Scene Content Diversity Analysis}

\begin{figure}[!t]
\vspace{-1mm}
    \centering
    \includegraphics[width=0.5\textwidth]{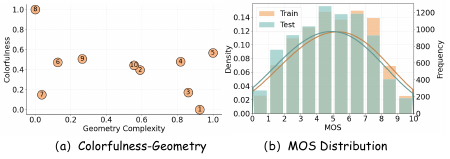}
    \vspace{-8mm}
    \caption{ MOS distribution for both training and testing viewpoints, with fitted density curves overlaid on the histogram. } 
    \label{fig:distribution}
    \vspace{-3mm}
\end{figure}

\begin{table*}[t]
\vspace{-3mm}
\caption{Performance benchmark on 3DGS-IEval-15K. $\spadesuit$ Handcrafted-based IQA models, $\blacklozenge$ Deep learning-based IQA models, $\heartsuit$ LLM Zero-Shot models.}
%The best results are marked in \textcolor{red}{RED} and the second-best in \textcolor{blue}{BLUE}.}
\renewcommand{\arraystretch}{0.8}
\centering
\vspace{-3mm}
\resizebox{1\textwidth}{!}{
  \begin{tabular}{l||ccc|ccc|ccc|ccc}
    \toprule
    \multicolumn{1}{l}{\bf Distortion Type} &
    \multicolumn{3}{c}{\textbf{All}} &
    \multicolumn{3}{c}{\textbf{Geometry-Only}} &
    \multicolumn{3}{c}{\textbf{Color-only}} &
    \multicolumn{3}{c}{\textbf{Geometry \& Color Mix}} \\
    \cmidrule(lr){2-4} \cmidrule(lr){5-7} \cmidrule(lr){8-10} \cmidrule(lr){11-13}
    \textbf{Methods / Metrics} & \textbf{SRCC} & \textbf{PLCC} & \textbf{KRCC} & \textbf{SRCC} & \textbf{PLCC} & \textbf{KRCC} & \textbf{SRCC} & \textbf{PLCC} & \textbf{KRCC} & \textbf{SRCC} & \textbf{PLCC} & \textbf{KRCC} \\
    \hline
    $\spadesuit$ PSNR~\citep{wang2004image} & 0.6451 & 0.6387 & 0.4574 & 0.4769 & 0.5141 & 0.3326 & 0.6808 & 0.6696 & 0.4863 & 0.6372 & 0.6493 & 0.4547 \\
    $\spadesuit$ SSIM~\citep{wang2004image} & 0.6790 & 0.6651 & 0.4889 & 0.6246 & 0.6448 & 0.4475 & 0.6803 & 0.6633 & 0.4875 & 0.6687 & 0.6542 & 0.4844 \\
    $\spadesuit$ MS-SSIM~\citep{wang2003multiscale} & 0.6983 & 0.6769 & 0.5061 & 0.6209 & 0.6411 & 0.4444 & 0.7021 & 0.6813 & 0.5078 & 0.6947 & 0.6682 & 0.5082 \\
    $\spadesuit$ IW-SSIM~\citep{wang2010information} & 0.6943 & 0.6918 & 0.5060 & 0.7068 & 0.7305 & 0.5187 & 0.6654 & 0.6635 & 0.4768 & 0.7007 & 0.6893 & 0.5142 \\
    $\spadesuit$ VIF~\citep{sheikh2006image} & 0.5590 & 0.5689 & 0.3928 & 0.5525 & 0.5644 & 0.3907 & 0.4913 & 0.5085 & 0.3400 & 0.5799 & 0.6062 & 0.4108 \\
    $\spadesuit$ FSIM~\citep{zhang2011fsim} & 0.7327 & 0.7124 & 0.5361 & 0.6922 & 0.7210 & 0.5058 & 0.7097 & 0.7052 & 0.5149 & 0.7305 & 0.7023 & 0.5366 \\
    $\spadesuit$ BRISQUE~\citep{mittal2012no} & 0.2202 & 0.2254 & 0.1495 & 0.3746 & 0.3734 & 0.2560 & 0.0739 & 0.0875 & 0.0498 & 0.2594 & 0.2591 & 0.1759 \\
    \midrule
    % $\heartsuit$ CLIP-IQA~\citep{wang2023exploring} & 0.3618 & 0.3357 & 0.2454 & 0.4573 & 0.4258 & 0.3096 & 0.2524 & 0.2174 & 0.1747 & 0.3798 & 0.3719 & 0.2531 \\
    $\heartsuit$ Llava-one-vision (0.5B)~\citep{xiong2024llavaovchat} & 0.2286 & 0.2311 & 0.1867 & 0.1075 & 0.1179 & 0.0876 & 0.2292 & 0.2190 & 0.1873 & 0.2180 & 0.2353 & 0.1783 \\
    $\heartsuit$ Llava-one-vision (7B)~\citep{xiong2024llavaovchat} & 0.4628 & 0.4821 & 0.3774 & 0.2690 & 0.3323 & 0.2204 & 0.3115 & 0.3527 & 0.2547 & 0.5197 & 0.5048 & 0.4232 \\
    $\heartsuit$ DeepSeekVL (7B)~\citep{lu2024deepseekvlrealworldvisionlanguageunderstanding} & 0.7065 & 0.6891 & 0.5480 & 0.6693 & 0.6621 & 0.5134 & 0.6800 & 0.6558 & 0.5127 & 0.6778 & 0.6855 & 0.5320 \\
    $\heartsuit$ LLaVA-1.5 (7B)~\citep{liu2024improvedbaselinesvisualinstruction} & 0.5754 & 0.5512 & 0.4495 & 0.3373 & 0.3660 & 0.2586 & 0.5513 & 0.5470 & 0.4247 & 0.5610 & 0.5553 & 0.4450 \\
    $\heartsuit$ LLaVA-NeXT (8B)~\citep{li2024llava} & 0.5899 & 0.5896 & 0.4791 & 0.5647 & 0.6024 & 0.4574 & 0.4435 & 0.4306 & 0.3525 & 0.5943 & 0.5918 & 0.4845 \\
    $\heartsuit$ mPLUG-Owl3 (7B)~\citep{ye2024mplugowl3longimagesequenceunderstanding} & 0.3575 & 0.3515 & 0.2687 & 0.5452 & 0.5616 & 0.4291 & 0.3626 & 0.3311 & 0.2711 & 0.2960 & 0.3009 & 0.2234 \\
    $\heartsuit$ Qwen2.5-VL (7B)~\citep{bai2025qwen25vltechnicalreport} & 0.7275 & 0.6985 & 0.5806 & 0.7114 & 0.6851 & 0.5656 & 0.6948 & 0.6350 & 0.5487 & 0.6992 & 0.6970 & 0.5606 \\
    $\heartsuit$ Llama3.2-Vision (11B)~\citep{meta_llama3_2024} & 0.0681 & 0.0032 & 0.0520 & 0.0940 & 0.0127 & 0.0740 & 0.0346 & 0.0122 & 0.0269 & 0.1464 & 0.0350 & 0.1150 \\
    $\heartsuit$ CogAgent (18B)~\citep{hong2024cogagentvisuallanguagemodel} & 0.5262 & 0.5100 & 0.4073 & 0.5452 & 0.5616 & 0.4291 & 0.4855 & 0.4609 & 0.3712 & 0.4665 & 0.4794 & 0.3632 \\
    $\heartsuit$ InternVL2.5 (8B)~\citep{chen2025expandingperformanceboundariesopensource} & 0.6698 & 0.0958 & 0.5317 & 0.6460 & 0.0843 & 0.5011 & 0.6389 & 0.1043 & 0.5109 & 0.6394 & 0.1105 & 0.5132 \\
    $\heartsuit$ InternVL3 (9B)~\citep{chen2025expandingperformanceboundariesopensource} & 0.5052 & 0.5032 & 0.3796 & 0.5901 & 0.5592 & 0.4420 & 0.4887 & 0.4672 & 0.3655 & 0.4824 & 0.5103 & 0.3666 \\
    
    $\heartsuit$ Gemini1.5-pro~\citep{google_gemini1.5_2024} & 0.6880 & 0.0893 & 0.5410 & 0.7125 & 0.1041 & 0.5589 & 0.6704 & 0.0189 & 0.5236 & 0.6653 & 0.1085 & 0.5309 \\
    $\heartsuit$ Q-Align~\citep{wu2023qalignteachinglmmsvisual}
    & 0.7711 & 0.7646 & 0.5668 & 0.7377 & 0.7413 & 0.5365 & 0.7250 & 0.7262 & 0.5164 & 0.7554 & 0.7657 & 0.5575 \\
    \midrule
    $\blacklozenge$ LPIPS (VGG)~\citep{zhang2018unreasonable} & 0.7313 & 0.7324 & 0.5367 & 0.7173 & 0.7152 & 0.5229 & 0.7087 & 0.7061 & 0.5119 & 0.7458 & 0.7589 & 0.5545 \\
    $\blacklozenge$ LPIPS (Alex)~\citep{zhang2018unreasonable} & 0.6767 & 0.6679 & 0.4870 & 0.6509 & 0.6675 & 0.4656 & 0.6732 & 0.6451 & 0.4824 & 0.7090 & 0.6967 & 0.5201 \\
    $\blacklozenge$ DISTS~\citep{ding2020image} & 0.8198 & 0.8132 & 0.6215 & 0.7593 & 0.7699 & 0.5613 & 0.7925 & 0.7912 & 0.5905 & 0.8342 & 0.8288 & 0.6428 \\
    $\blacklozenge$ {TReS}~\citep{golestaneh2022noreferenceimagequalityassessment} 
    & 0.7969 & 0.7966 & 0.5988 & 0.6578 & 0.6891 & 0.4749 & 0.7496 & 0.7748 & 0.5529 & 0.7912 & 0.8005 & 0.5988 \\
    $\blacklozenge$ {DBCNN}~\citep{Zhang_2020}  %\cite{he2016deep} 
    & 0.8635 & 0.8551 & 0.6699 & 0.7770 & 0.7896 & 0.5817 & 0.8367 & 0.8350 & 0.6414 & 0.8471 & 0.8536 & 0.6578\\
    $\blacklozenge$ {MANIQA}~\citep{yang2022maniqamultidimensionattentionnetwork} 
    & 0.9356 & 0.9338 & 0.7786 & 0.8443 & 0.8727 & 0.6623 & 0.8999 & 0.9160 & 0.7338 & 0.9277 & 0.9323 & 0.7717 \\
    $\blacklozenge$ {STAIRIQA}~\citep{Sun_2023} 
    & 0.9348 & 0.9325 & 0.7769 & 0.8596 & 0.8795 & 0.6781 & 0.9016 & 0.9150 & 0.7327 & 0.9236 & 0.9292 & 0.7657 \\
    $\blacklozenge$ {MUSIQ}~\citep{ke2021musiqmultiscaleimagequality} 
    & 0.9308 & 0.9287 & 0.7697 & 0.8454 & 0.8705 & 0.6624 & 0.8934 & 0.9089 & 0.7220 & 0.9200 & 0.9270 & 0.7606 \\
    $\blacklozenge$ {HYPERIQA}~\citep{su2020blindly} 
    & 0.9407 & 0.9391 & 0.8121 & 0.8785 & 0.8981 & 0.7171 & 0.9086 & 0.9199 & 0.7626 & 0.9382 & 0.9385 & 0.8120 \\
    $\blacklozenge$ {LIQE}~\citep{zhang2023blind} 
    & 0.9190 & 0.9043 & 0.7496 & 0.8135 & 0.8364 & 0.6251 & 0.8816 & 0.8921 & 0.7010  & 0.9176 & 0.9011 & 0.7531 \\

    \bottomrule
  \end{tabular}
}
\label{benchmark}
\vspace{-3mm}
\end{table*}

To validate selected source scene diversity, we measure geometry and color complexity using spatial perceptual information (SI) ~\citep{ITU1(siti)} and colorfulness metrics (CM) ~\citep{ITU2}, respectively. As shown in Figure~\ref{fig:distribution} (a), where each numbered point corresponds to the scene index in Figure~\ref{figure:model}, the uniform distribution across both complexity dimensions confirms our dataset covers diverse visual characteristics, ensuring comprehensive evaluation of 3DGS methods across varying challenging scenarios.

\subsection{MOS and Viewpoint-Based Quality Analysis}

\textbf{MOS Distribution:} As presented in Figure~\ref{fig:distribution} (b), our dataset demonstrates comprehensive quality coverage across the entire 0-10 MOS range with sufficient samples in each score segment, ensuring adequate representation of varying distortion levels, and providing a robust foundation for training quality assessment models.

\textbf{Inter-View Quality Disparity:} As shown in Figure~\ref{fig:distribution}(b), test viewpoints exhibit lower MOS scores than training viewpoints, with distributions peaking around 4-5 versus 5-6 respectively. Specifically, for MOS scores $\le$ 6, novel views contribute more samples, while for MOS > 6, training views outnumber test views.

This represents the first systematic investigation of quality differences between training and novel viewpoints in 3DGS, revealing substantial view-dependent quality variation that provides insights for future 3DGS optimization and quality assessment design.
\vspace{-2mm}
\subsection{Performance on Score Prediction}

\subsubsection{Experiment Settings} \

\textbf{Evaluation Metrics.} To evaluate the correlation between the predicted scores and the ground-truth MOSs, we employ three widely used evaluation criteria: Spearman Rank-order Correlation Coefficient (SRCC), Pearson Linear Correlation Coefficient (PLCC), and Kendall Rank Correlation Coefficient (KRCC).

\textbf{Reference Algorithms.} We comprehensively evaluate 30 representative image quality assessment algorithms across three categories: handcrafted-based IQA models, LLM Zero-Shot models, and deep learning-based IQA models.

\textbf{Dataset Partitioning.} We construct four dataset configurations to isolate distortion effects: \textbf{Geometry-Only} (purely geometric distortions), \textbf{Color-Only} (color-related degradations only), \textbf{Geometry \& Color Mix} (simultaneous geometric and color distortions).
 For each of these three subsets, we apply a 4:1 train-test split across all scenes. Based on these subsets, we further construct the \textbf{All} setting by concatenating the training sets and testing sets from the three subsets, respectively, thus also preserving the 4:1 training-to-testing ratio. This setup enables a comprehensive evaluation across individual and combined distortion types while ensuring consistent data partitioning. 

\textbf{Training Settings.} Traditional handcrafted metrics are directly evaluated on corresponding datasets. LLM Zero-Shot models use pre-trained weights for inference. Deep learning-based IQA models are trained only on the training set of the All configuration and evaluate directly on the test sets of all four configurations without any fine-tuning on the individual subsets. This experimental design enables us to examine the same model's performance across different distortion scenarios, thereby comprehensively exploring quality assessment methods' generalization capabilities across diverse distortion types.

 \vspace{-5pt}
\subsubsection{Results and Analysis} \

The results reveal distinct performance patterns across the three model categories. Deep learning-based IQA models achieve the highest performance, with top methods like HYPERIQA and MANIQA reaching SRCC values exceeding 0.93 on the All dataset. Handcrafted IQA models demonstrate moderate performance, with FSIM achieving the best SRCC of 0.7327, while traditional metrics like PSNR and SSIM show substantially lower correlations around 0.64-0.68. LLM Zero-Shot models exhibit the most varied performance, ranging from near-zero correlations (Llama3.2-Vision: 0.0681) to competitive results (Q-Align: 0.7711), though notably, these models were not fine-tuned for the quality assessment task. The superior performance of deep learning-based methods stems from their learned perceptual representations that better capture human visual perception, while handcrafted metrics rely on predetermined mathematical formulations that may not align with human judgment. The variable performance of LLM Zero-Shot models reflects their primary design for general visual understanding rather than specialized quality assessment, though their semantic reasoning capabilities show promise for this domain.

Examining performance across distortion-specific evaluations reveals a consistent pattern: most methods exhibit performance degradation when evaluated on isolated distortion types compared to the comprehensive All dataset. For instance, MANIQA's SRCC drops from 0.9356 (All) to 0.8443 (Geometry-Only) and 0.8999 (Color-Only). Similarly, HYPERIQA shows a decline from 0.9407 (All) to 0.8785 (Geometry-Only) and 0.9086 (Color-Only). This phenomenon indicates that while these models achieve strong overall performance, they struggle with domain-specific distortions that differ from their training distribution. The performance gaps suggest that models benefit from the diverse distortion patterns present in the All training set, and their generalization to isolated distortion types remains challenging, highlighting the importance of distortion-specific evaluation for comprehensive model assessment.

\section{Conclusion}
In this paper, we present 3DGS-IEval-15K, a comprehensive dataset designed to advance image quality assessment for compressed 3DGS representations. It provides 15,200 images from 10 real-world scenes using 6 mainstream 3DGS compression algorithms across 20 strategically selected viewpoints, incorporating systematic compression parameter design and controlled subjective experiments with 60 viewers. Moreover, we establish a comprehensive benchmark of 30 representative IQA metrics, including deep learning-based and LLM-based approaches on 3DGS dataset for the first time. We demonstrate view-dependent quality variations for 3DGS. We reveal limitations and make suggestions of existing metrics in capturing 3DGS-specific compression distortions. 3DGS-IEval-15K aims to support the development of specialized quality assessment models and provide insights for optimizing compression techniques aligned with human visual perception, ultimately advancing 3DGS toward practical deployment.

\section*{Acknowledgments}
This paper is supported in part by National Natural Science Foundation of China (62371290, U20A20185), the Fundamental Research Funds for the Central Universities of China, and STCSM under Grant (22DZ2229005). The corresponding author is Yiling Xu(e-mail: yl.xu@sjtu.edu.cn).

% \begin{acks}
% \end{acks}

\bibliographystyle{ACM-Reference-Format}
\bibliography{main}

% \newpage
% \appendix
% \twocolumn[{\section*{\centering LMME3DHF: Benchmarking and Evaluating Multimodal 3D \\ Human Face Generation with LMMs (Supplemental Materials)\\}}\vspace*{10mm}]
% \input{appendix/0_overview}
% \input{appendix/1_generation_model}
% \input{appendix/2_subjective_experiment}
% \input{appendix/3_database}
% \input{appendix/4_train_loss}
% \input{appendix/5_implementions}

% \input{appendix/figures_tables/sal_map_example}
% \input{appendix/figures_tables/saliency_predict_example}

\end{document}